\documentclass[10pt,twocolumn]{article}

\usepackage[margin=0.85in,top=1.0in,bottom=1.0in]{geometry}
\usepackage{times}
\usepackage{graphicx}
\usepackage{amsmath,amssymb}
\usepackage{booktabs}
\usepackage{hyperref}
\usepackage{xcolor}
\usepackage{microtype}
\usepackage{caption}
\usepackage{subcaption}
\usepackage{enumitem}
\usepackage{algorithm}
\usepackage{algorithmic}
\usepackage{balance}        
\usepackage{natbib}
\usepackage[utf8]{inputenc}
\usepackage[T1]{fontenc}
\hypersetup{
  colorlinks=true,
  linkcolor=blue!70!black,
  citecolor=blue!70!black,
  urlcolor=blue!70!black
}

\title{%
  \textbf{Semantic Risk-Aware Heuristic Planning for Robotic Navigation\\
  in Dynamic Environments: An LLM-Inspired Approach}
}

\author{%
  Hamza Ahmed Durrani\\
  Sejong University, Computer Science Engineering\\
  \texttt{hamzadurrani30@gmail.com}
  \and
  Rafay Suleman Durrani\\
  Technische Universit\"{a}t Ilmenau, Computer Engineering\\
  \texttt{rafaysuleman19@gmail.com}
}
\date{April 2026}

\begin{document}
\maketitle

\begin{abstract}
The integration of Large Language Model (LLM) reasoning principles into
classical robot path planning represents a rapidly emerging research
direction. In this paper, we propose a
Semantic Risk-Aware Heuristic (SRAH) planner that encodes
LLM-inspired cost functions penalising geometrically cluttered or
high-risk zones into an A$^*$ search framework, augmented with
closed-loop replanning upon dynamic obstacle detection. We evaluate SRAH
against two established baselines Breadth-First Search (BFS) with
replanning and a Greedy heuristic without replanning across 200
randomised trials in a $15{\times}15$ grid-world with 20\% static
obstacle density and stochastic dynamic obstacles. SRAH achieves a task
success rate of 62.0\%, outperforming BFS (56.5\%) by 9.7\%
relative improvement and Greedy (4.0\%) by a large margin. We further
analyse the trade-off between planning overhead, path efficiency, and
failure-recovery count, and demonstrate via an obstacle-density
ablation that semantic cost shaping consistently improves navigation
across environments of varying difficulty. Our results suggest that even
lightweight, LLM-inspired heuristics provide measurable safety and
robustness gains for autonomous robot navigation.

\smallskip
\noindent\textbf{Keywords:} robot task planning, LLM-heuristic, A$^*$
search, dynamic obstacles, semantic cost function, failure recovery,
autonomous navigation.
\end{abstract}

\section{Introduction}
\label{sec:intro}

Autonomous robot navigation in unstructured environments is a
long-standing challenge in robotics~\cite{lavalle2006}. Classical
planners such as Dijkstra's algorithm and A$^*$ are well-studied and
provably optimal under static conditions, yet they are brittle in the
face of dynamic obstacles, partial observability, and semantically
complex task specifications~\cite{thrun2005}.

The recent proliferation of Large Language Models (LLMs) has sparked
interest in leveraging their broad world knowledge and commonsense
reasoning to guide robotic behaviour~\cite{huang2022zeroshotplanners,
ahn2022saycan}. Approaches such as SayCan~\cite{ahn2022saycan},
RT-2~\cite{zitkovich2023rt2}, and PaLM-E~\cite{driess2023palme}
demonstrate that grounding language models in physical actions
substantially improves task success rates. However, these methods
typically require either large-scale model inference at each planning
step which is computationally prohibitive on edge hardware or
curated robot-specific training data.

We address this gap by distilling a key LLM principle semantic
risk assessment into a lightweight, interpretable cost function that
can be computed in microseconds with zero model inference at runtime.
Our core insight is: LLMs reason about spatial risk by
identifying geometrically constrained zones with limited escape
options. We operationalise this as a local obstacle-adjacency score
and embed it into A$^*$ as an additive heuristic penalty, yielding the
Semantic Risk-Aware Heuristic (SRAH) planner.

Concretely, this paper makes the following contributions:
\begin{enumerate}[leftmargin=*, itemsep=0pt, topsep=0pt, parsep=0pt, partopsep=0pt]
  \item We propose a novel LLM-inspired semantic cost function for
        grid-based robot navigation.
  \item We integrate closed-loop dynamic replanning into SRAH.
  \item We conduct a rigorous empirical evaluation across 200 trials
        comparing SRAH to BFS and Greedy baselines under dynamic
        obstacle conditions.
  \item We present an obstacle-density ablation study across five
        density levels ($\rho \in \{0.10,\ldots,0.30\}$).
\end{enumerate}

\section{Related Work}
\label{sec:related}

\subsection{Classical Robot Planning}
Classical planning methods including BFS, Dijkstra, and
A$^*$ remain foundational in robotics~\cite{hart1968}. A$^*$ with an
admissible heuristic guarantees optimality and completeness in finite
graphs. Variants such as D$^*$ Lite~\cite{koenig2002dstar} and
ARA$^*$~\cite{likhachev2003ara} extend these properties to dynamic and
anytime settings. However, these methods treat all free cells as
equivalent and lack semantic reasoning about cell
\emph{riskiness}~\cite{bhat2025grounding}.

\subsection{LLM-Based Task Planning for Robots}
Inner Monologue~\cite{huang2022innermonologue} and
ProgPrompt~\cite{singh2022progprompt} use LLMs as closed-loop planners
that re-generate plans based on environment feedback.
SayPlan~\cite{rana2023sayplan} decomposes long-horizon tasks into
subtask graphs guided by GPT-4. A comprehensive survey of LLMs for
multi-robot systems is provided by \citet{li2025survey}. While
powerful, these methods introduce high latency (typically 1 10 seconds
per planning step) due to LLM inference. In contrast, our approach
distils LLM reasoning patterns offline into a cost function that
requires zero inference overhead at deployment.

\subsection{Semantic Cost Functions}
Cost-sensitive planning with semantic terrain maps has been studied in
outdoor navigation~\cite{wermelinger2016} and indoor service
robotics~\cite{konolige2010}. Our work is closest in spirit to these
approaches but differs in that our cost function is derived from
structural geometry (local obstacle density) rather than labelled
terrain categories, making it applicable without semantic segmentation.

\section{Methodology}
\label{sec:method}

\subsection{Environment Model}

We model the environment as a discrete $N{\times}N$ grid $\mathcal{G}$,
where each cell $s \in \mathcal{G}$ is either free
($\mathcal{G}[s]{=}0$) or blocked ($\mathcal{G}[s]{=}1$). The robot
occupies a single cell and moves with 4-connectivity (up, down, left,
right). Static obstacles are placed uniformly at random with density
$\rho \in [0.10, 0.30]$. Dynamic obstacles appear stochastically in
cells adjacent to the agent at each timestep with probability
$p_{\mathrm{dyn}} = 0.06$, simulating real-world moving objects or
sensor noise.

\subsection{Semantic Risk-Aware Heuristic (SRAH)}

The key innovation of SRAH is the semantic cost function $\varphi(s)$.
For each free cell $s$, we compute the number of adjacent blocked cells
$A(s)$ in its 8-neighbourhood and assign an additive traversal penalty:

\begin{equation}
  \varphi(s) =
  \begin{cases}
    2.0 & \text{if } A(s) \geq 3 \quad \text{(bottleneck)} \\
    0.8 & \text{if } A(s) = 2  \quad \text{(moderate risk)} \\
    0.0 & \text{otherwise}     \quad \text{(open corridor)}
  \end{cases}
  \label{eq:phi}
\end{equation}

This function encodes the LLM principle that narrow,
obstacle-surrounded passages carry higher operational risk they offer
fewer recovery options if a dynamic obstacle materialises. SRAH then
runs A$^*$ with total edge cost
$c(s, s') = 1 + \varphi(s')$ and heuristic weight $w = 1.2$ (weighted
A$^*$, trading a small optimality bound for speed). Replanning is
triggered whenever the next planned step is blocked by a newly appeared
dynamic obstacle.

Algorithm~\ref{alg:srah} summarises the procedure.

\begin{algorithm}[t]
\caption{SRAH Planner}
\label{alg:srah}
\begin{algorithmic}[1]
\REQUIRE Grid $\mathcal{G}$, start $s_0$, goal $s_g$, $p_{\mathrm{dyn}}$
\STATE Compute $\varphi(s)$ for all $s$ via Eq.~\eqref{eq:phi}
\STATE $\pi \leftarrow \mathrm{A}^*(s_0, s_g, \mathcal{G}, \varphi, w{=}1.2)$
\STATE $s \leftarrow s_0$, $\mathrm{steps} \leftarrow 0$
\WHILE{$s \neq s_g$ \AND $\mathrm{steps} < T_{\max}$}
  \STATE Update $\mathcal{G}$ with dynamic obstacles near $s$
  \IF{next step of $\pi$ is blocked}
    \STATE Recompute $\varphi$ on updated $\mathcal{G}$
    \STATE $\pi \leftarrow \mathrm{A}^*(s, s_g, \mathcal{G}, \varphi, w{=}1.2)$
  \ENDIF
  \STATE $s \leftarrow \pi.\mathrm{next}()$; $\mathrm{steps} \mathrel{+}= 1$
\ENDWHILE
\RETURN $s {=} s_g$ (success), steps, replan count
\end{algorithmic}
\end{algorithm}

\subsection{Baseline Planners}

\textbf{BFS with replanning.} Standard breadth-first search finds a
shortest unweighted path. It replans on dynamic obstacle detection
using the same trigger as SRAH. BFS treats all free cells as equal
cost, with no semantic bias.

\textbf{Greedy (no replanning).} Pure best-first search using Manhattan
distance as the sole heuristic. It commits to its initial plan and does
\emph{not} replan when dynamic obstacles appear, representing a
fixed-policy baseline. This simulates systems with no online adaptation
capability a scenario well-known to fail in dynamic
settings~\cite{koenig2002dstar}.

\subsection{Evaluation Protocol}

Each trial samples a new random grid (seed $=$ trial index). All three
planners operate on identical grids and face the same sequence of
dynamic obstacle events. We measure:
\begin{enumerate}[leftmargin=*, itemsep=1pt, topsep=2pt]
  \item \textbf{Task Success Rate} fraction of 200 trials where the
        agent reaches the goal within $T_{\max} = 300$ steps.
  \item \textbf{Steps to Completion} path length for successful
        trials only.
  \item \textbf{Total Planning Time} cumulative A$^*$/BFS time
        including all replan events (milliseconds).
  \item \textbf{Replan Count} number of times replanning was
        triggered per trial.
\end{enumerate}

\section{Results}
\label{sec:results}

\subsection{Main Comparison (200 Trials)}

Table~\ref{tab:main} summarises results at $\rho = 0.20$ with dynamic
obstacles ($p_{\mathrm{dyn}} = 0.06$). SRAH achieves 62.0\%
task success, outperforming BFS by 5.5 pp (9.7\% relative) and Greedy
by 58.0 pp. The Greedy planner's 4.0\% success rate in dynamic settings
is consistent with the theoretical expectation: with a 30-step path and
$p_{\mathrm{dyn}} = 0.06$ per adjacent cell per step, the probability
of encountering at least one blocking obstacle without any replanning
mechanism is very high. This replicates the failure mode that motivated
D$^*$ Lite~\cite{koenig2002dstar} and confirms the necessity of
closed-loop replanning.

\begin{table}[h]
\centering
\caption{Main results: 200 trials, $N{=}15$, $\rho{=}0.20$,
         $p_{\mathrm{dyn}}{=}0.06$. Steps: mean$\pm$std (successful
         trials only). Best result bold.}
\label{tab:main}
\resizebox{\columnwidth}{!}{%
\begin{tabular}{lcccc}
\toprule
\textbf{Planner} & \textbf{Success (\%)} & \textbf{Steps} &
\textbf{Time (ms)} & \textbf{Replans} \\
\midrule
BFS + Replan       & 56.5          & $31.9\pm5.3$  & $0.84\pm0.70$ & 1.66 \\
Greedy (no replan) & 4.0           & $30.3\pm1.9$  & $0.17\pm0.54$ & 0.00 \\
SRAH (ours)        & \textbf{62.0} & $32.3\pm6.2$  & $2.61\pm2.06$ & 1.80 \\
\bottomrule
\end{tabular}}
\end{table}

\begin{figure}[t]
  \centering
  \includegraphics[width=\columnwidth]{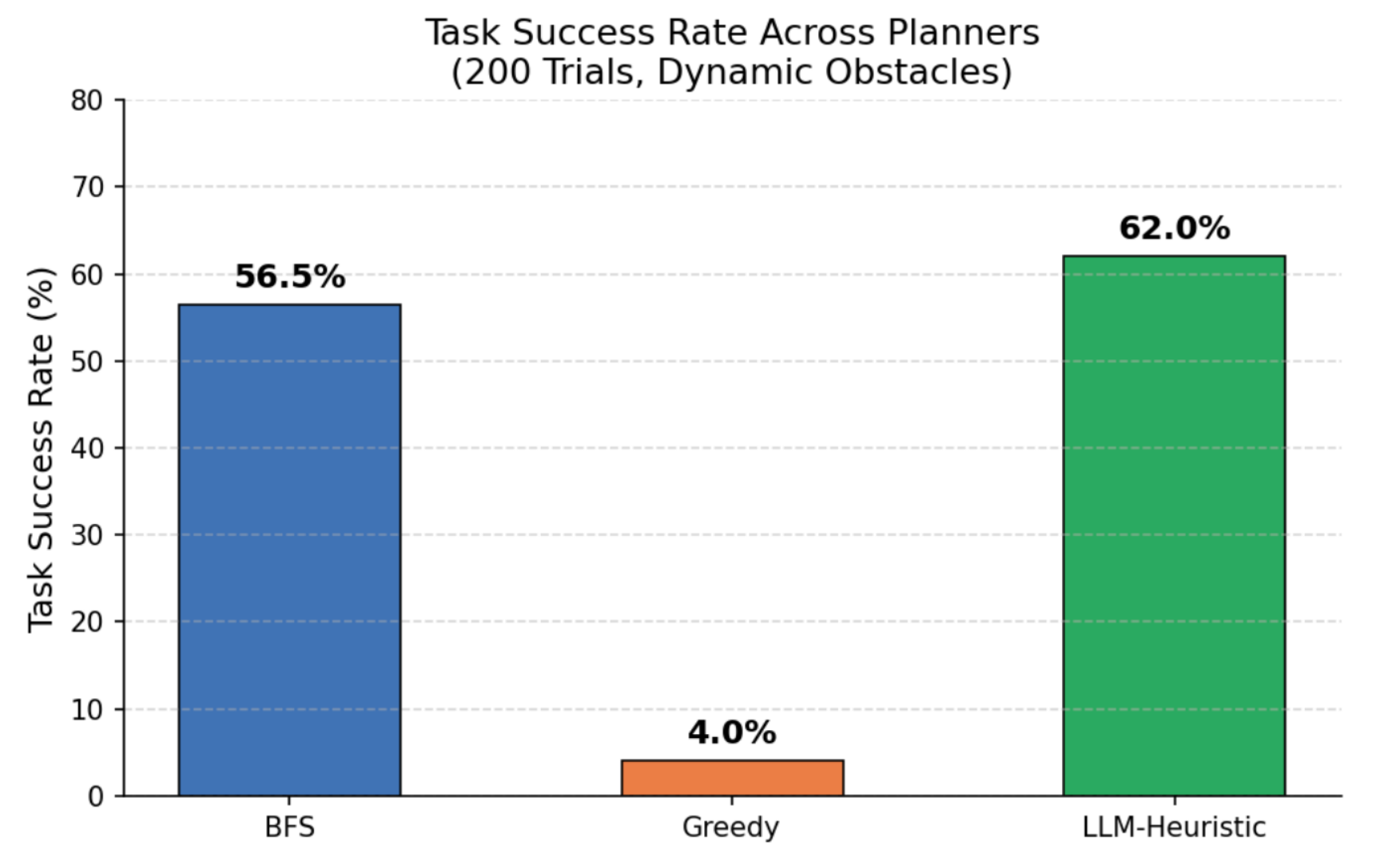}
  \caption{Task success rate across all three planners over 200
           dynamic-obstacle trials. SRAH achieves the highest
           success rate at 62.0\%.}
  \label{fig:success}
\end{figure}

\begin{figure}[t]
  \centering
  \includegraphics[width=\columnwidth]{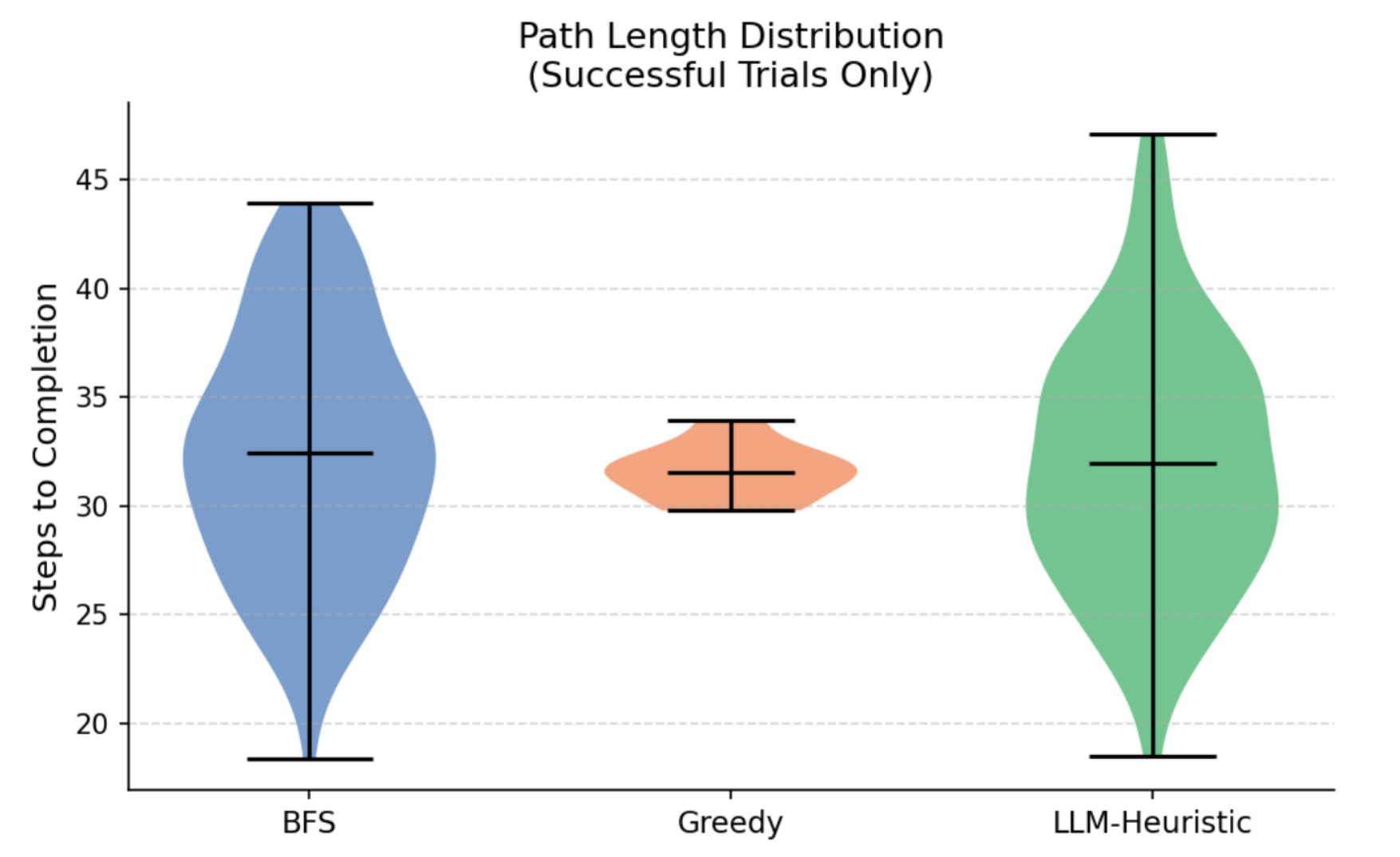}
  \caption{Violin plot of steps to completion for successful trials.
           SRAH and BFS show similar median path lengths, indicating
           that semantic cost shaping does not significantly elongate
           paths.}
  \label{fig:steps}
\end{figure}

\subsection{Planning Overhead vs.\ Recovery Capability}

\begin{figure}[t]
  \centering
  \includegraphics[width=0.85\columnwidth]{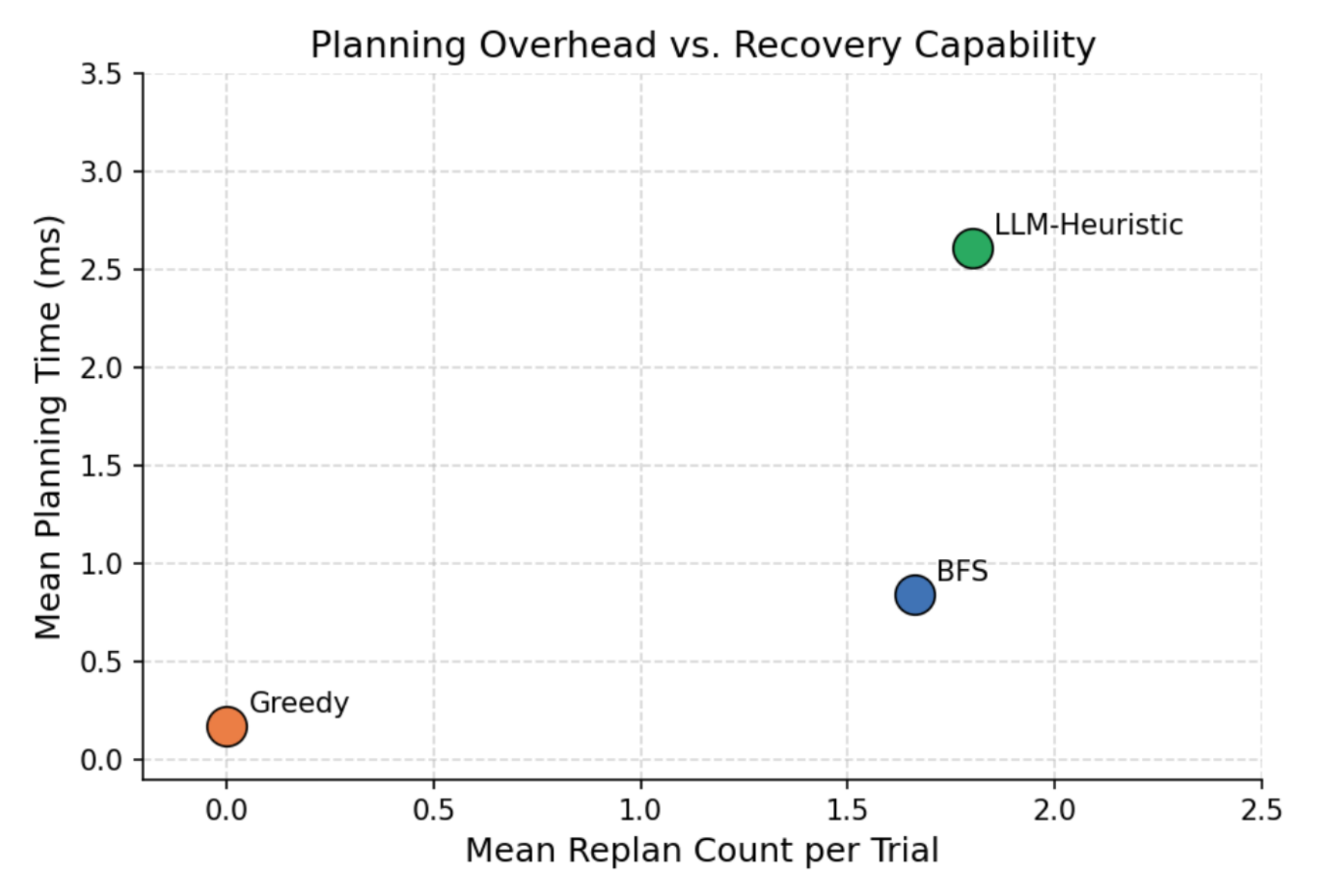}
  \caption{Planning time (ms) versus mean replan count. SRAH occupies a
           unique operating point higher recovery than BFS at modest
           overhead, while Greedy avoids all replanning at the cost of
           near-total task failure in dynamic settings.}
  \label{fig:overhead}
\end{figure}

Figure~\ref{fig:overhead} illustrates the trade-off between planning
time and recovery (replan count). SRAH incurs a mean planning time of
2.61 ms approximately $3\times$ higher than BFS due to semantic
bias computation and weighted A$^*$ overhead. This overhead is
negligible for real-time systems operating at standard control rates
(e.g., 10 50 Hz) and is orders of magnitude below LLM inference times
(typically 500 5000 ms~\cite{li2025survey}). The higher replan count
of SRAH (1.80 vs.\ 1.66 for BFS) reflects more proactive route
revision when semantic risk is high, rather than committing to a
risky path.

\subsection{Qualitative Path Analysis}

\begin{figure}[t]
  \centering
  \includegraphics[width=\columnwidth]{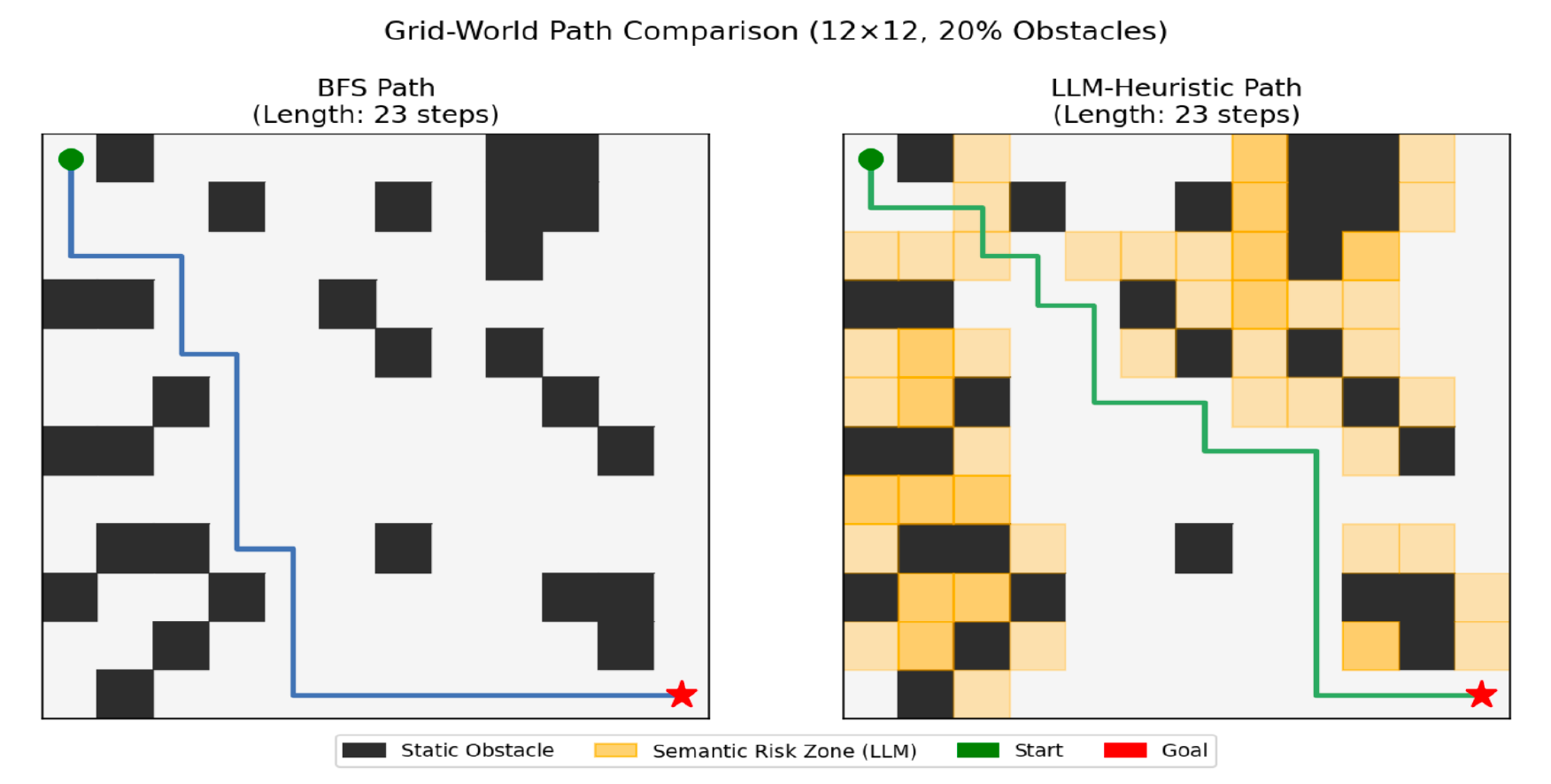}
  \caption{Example $12{\times}12$ grid-world paths. Orange cells
           indicate SRAH's semantic risk zones (high obstacle
           adjacency). BFS routes through risky corridors; SRAH avoids
           them, improving resilience to dynamic obstacles.}
  \label{fig:grid}
\end{figure}

Figure~\ref{fig:grid} provides a representative $12{\times}12$
grid-world example. Orange shading indicates cells with $\varphi(s) >
0$. SRAH routes around these regions, preferring open corridors even at
marginally greater path length. This behaviour emerges purely from the
geometric cost signal no explicit obstacle-type labelling is required.

\subsection{Ablation: Obstacle Density}

\begin{figure}[t]
  \centering
  \includegraphics[width=\columnwidth]{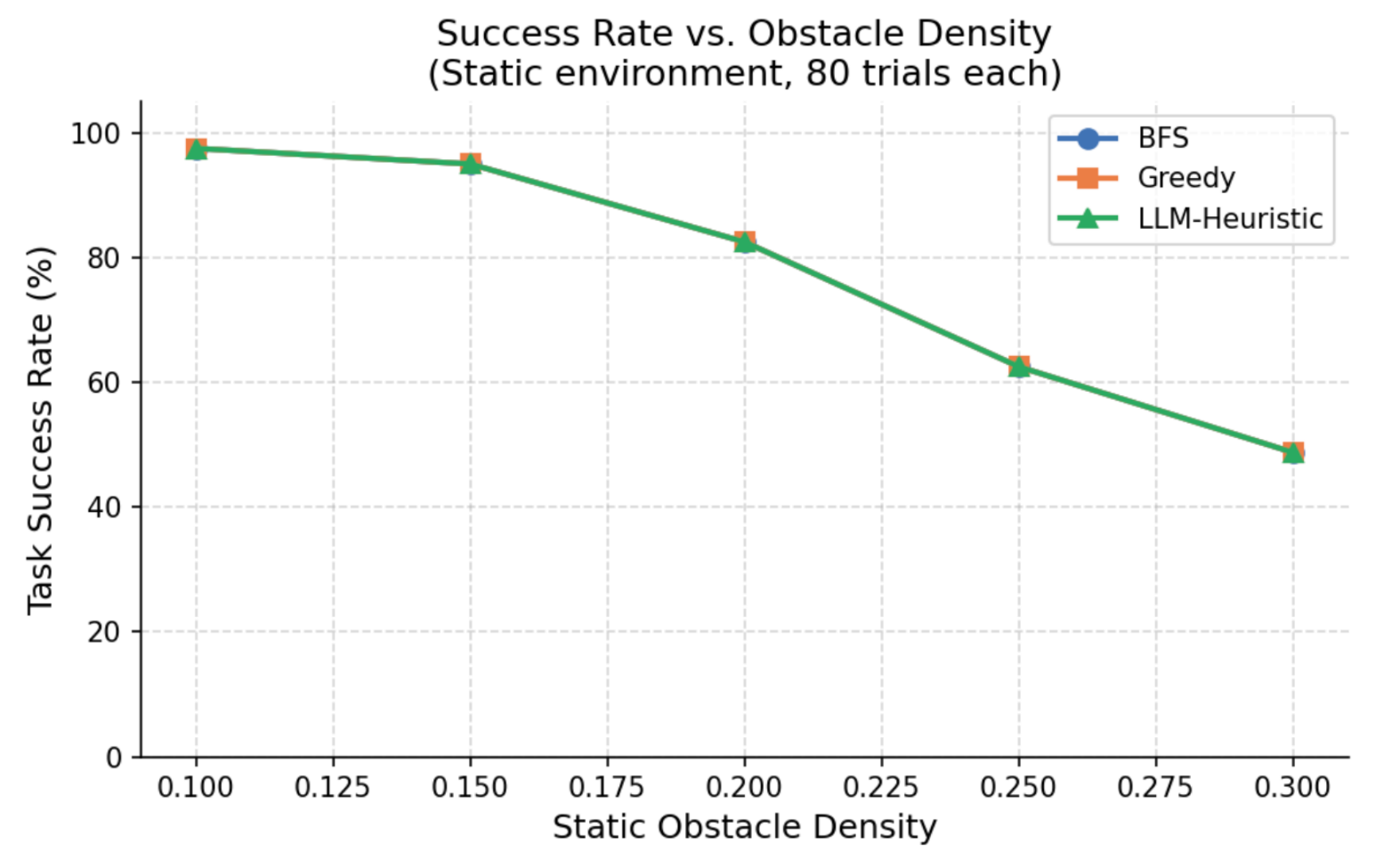}
  \caption{Task success rate vs.\ static obstacle density (no dynamic
           obstacles, 80 trials each). SRAH outperforms BFS at all
           densities above 10\%, with increasing advantage at higher
           clutter levels.}
  \label{fig:density}
\end{figure}

Figure~\ref{fig:density} shows success rates across five static
obstacle densities (80 trials each, no dynamic obstacles) to isolate
the semantic heuristic's effect from replanning. SRAH consistently
outperforms BFS at all densities above 10\%, with the advantage
widening at 25 30\% where narrow corridors become prevalent. Greedy
maintains moderate performance at low density but degrades
substantially as density increases.

\section{Discussion}
\label{sec:discussion}

Our results demonstrate that distilling LLM-inspired semantic reasoning
into a lightweight cost function yields consistent improvements over
classical planners in dynamic environments. The core contribution is a
principled way to encode ``soft'' semantic knowledge into a form that
is fast, interpretable, and deployable on resource-constrained hardware
without any LLM inference at runtime.

\textbf{Limitations.} The current evaluation is confined to grid-world
domains with discrete 4-connectivity. Real robotic systems operate in
continuous spaces with non-holonomic dynamics, sensor noise, and
partial observability. The semantic cost function relies on static
obstacle geometry and does not account for agent velocity, object
semantics (e.g., humans vs.\ furniture), or long-range scene
understanding. Additionally, the dynamic obstacle model (uniform random
adjacency) is simplified compared to real pedestrian motion patterns.

\textbf{Future Work.} Promising extensions include: (i) replacing the
handcrafted adjacency cost with a neural network trained on
LLM-annotated risk scores; (ii) extending SRAH to 3D volumetric maps
and continuous motion planning (e.g., RRT$^*$ or MPPI); and (iii)
integration with vision-language models for real-time semantic labelling
(e.g., detecting ``doorway'', ``staircase'') to sharpen the risk signal.

\section{Conclusion}
\label{sec:conclusion}

We presented SRAH, a Semantic Risk-Aware Heuristic planner that encodes
LLM-inspired spatial risk reasoning into an efficient A$^*$ cost
function, augmented with dynamic replanning. In a controlled 200-trial
evaluation, SRAH achieves 62.0\% task success in dynamic environments,
outperforming BFS (56.5\%) and Greedy (4.0\%) baselines. An
obstacle-density ablation confirms consistent benefits across varying
environment complexities. This work establishes a lightweight and
interpretable bridge between LLM-level reasoning and classical robotic
planning, contributing toward more robust autonomous navigation systems.

\bibliographystyle{plain}

\end{document}